\documentclass[letterpaper, 10pt, conference]{IEEEtran}
\IEEEoverridecommandlockouts
\usepackage{cite}
\usepackage{amsmath,amssymb,amsfonts}
\usepackage{algorithmic}
\usepackage{graphicx}
\usepackage{textcomp}
\usepackage{xcolor}
\usepackage{hyperref}
\usepackage[letterpaper, left=0.75in, right=0.75in, bottom=0.75in, top=0.75in]
{geometry}
\begin{document}

\title{\vspace{0.18in}PHGNet: Prototype-Guided Hypergraph Construction for Heterogeneous Spatiotemporal Forecasting\\
}

\author{
    \IEEEauthorblockN{%
    Ruiwen Gu\textsuperscript{1}, Yahao Liu\textsuperscript{2}, Zhenyu Liu\textsuperscript{1}, Qitai Tan\textsuperscript{1}, Xiao-Ping Zhang\textsuperscript{1,*}
}
    \IEEEauthorblockA{\textsuperscript{1} Shenzhen Ubiquitous Data Enabling Key Lab \\
    Shenzhen International Graduate School, Tsinghua University, Shenzhen, China}
    \IEEEauthorblockA{\textsuperscript{2} School of Computer Science and Engineering \\ 
    University of Electronic Science and Technology of China, Chengdu, China}
    \IEEEauthorblockA{\textsuperscript{1} \{grw23, tqt24\}@mails.tsinghua.edu.cn, zhenyuliu@sz.tsinghua.edu.cn, xpzhang@ieee.org\\ \textsuperscript{2} lyhaolive@gmail.com}
    \thanks{* Corresponding author.}
}

\maketitle
\begin{abstract}
As a core task in intelligent transportation systems, traffic forecasting plays a critical role in urban traffic management. Accurate traffic forecasting relies on modeling complex spatiotemporal dependencies, which is inherently challenging due to spatial heterogeneity in traffic systems.
Despite significant progress, most existing methods are still limited to pairwise spatial dependency modeling, making it difficult to capture dynamic high-order interactions among nodes with similar traffic patterns. To address this issue, we propose PHGNet, a novel spatiotemporal forecasting framework based on prototype-guided hypergraph construction. At the core of PHGNet, a prototype learning mechanism is designed to adaptively assign pattern-similar nodes to hyperedges, thereby capturing high-order interactions with time-varying structures. To improve the reliability of dynamic hypergraph construction, we further develop a global-local node representation module to extract time-consistent features. For forecasting, iterative residual refinement and Temporal Query Attention are introduced to improve forecasting accuracy while supporting efficient parallel decoding. Extensive experiments on multiple real-world datasets demonstrate that PHGNet achieves superior predictive performance compared with state-of-the-art methods.
\end{abstract}

\begin{IEEEkeywords}
Traffic prediction, hypergraph convolution, spatiotemporal data
\end{IEEEkeywords}

\section{Introduction}
Traffic forecasting is a fundamental task in intelligent transportation systems, aiming to predict future traffic conditions from historical time series collected by road sensors. Accurate traffic forecasting supports a wide range of practical applications, including route planning, signal control, and urban traffic management. With the rapid development of deep learning, neural network-based approaches have been extensively explored to address this challenging issue. Among them, spatiotemporal learning frameworks\cite{STGCN,DCRNN,GWNet,MTGNN,ADMFormer,STDN,PGCN} have become the dominant paradigm for traffic forecasting. In general, these methods model traffic dynamics as a diffusion process, typically integrating graph convolutions with temporal modules to capture spatiotemporal dependencies.

\begin{figure}[htbp]
    \centering
    \includegraphics[width=\columnwidth]{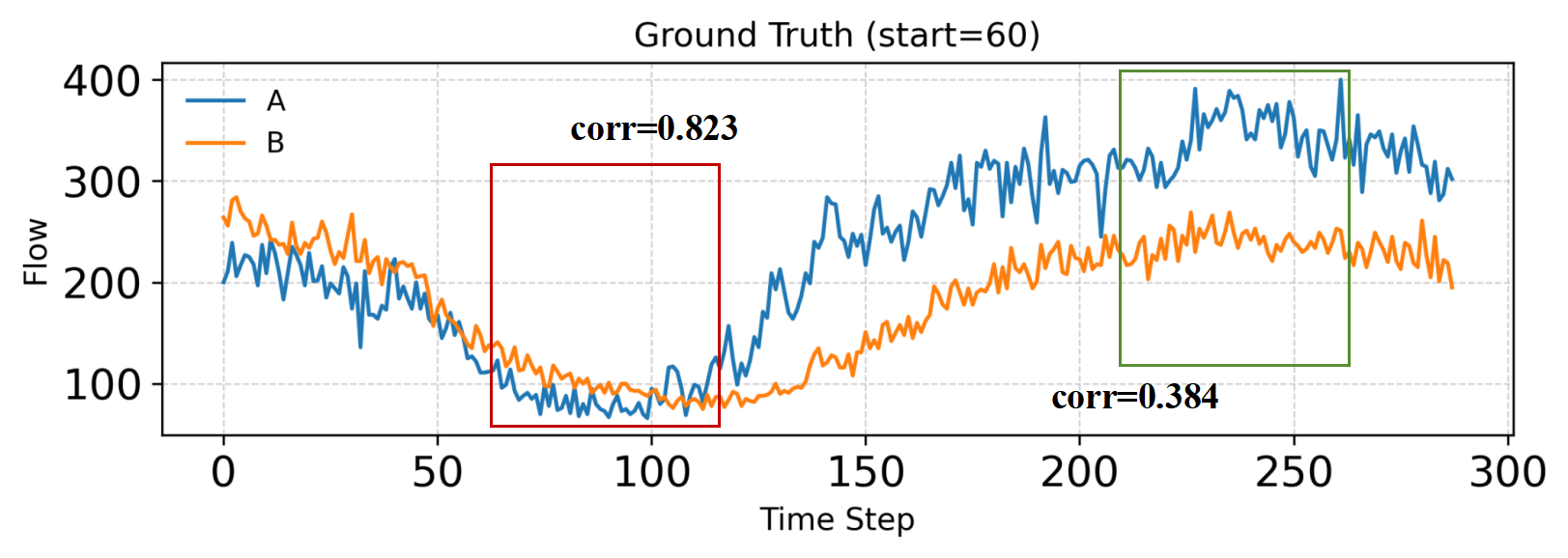} 
    \caption{Illustration of spatial heterogeneity in traffic flow}
    \label{fig:spatial hetero}
\end{figure}

\begin{figure*}[!t]
    \centering
    \includegraphics[width=0.85\textwidth]{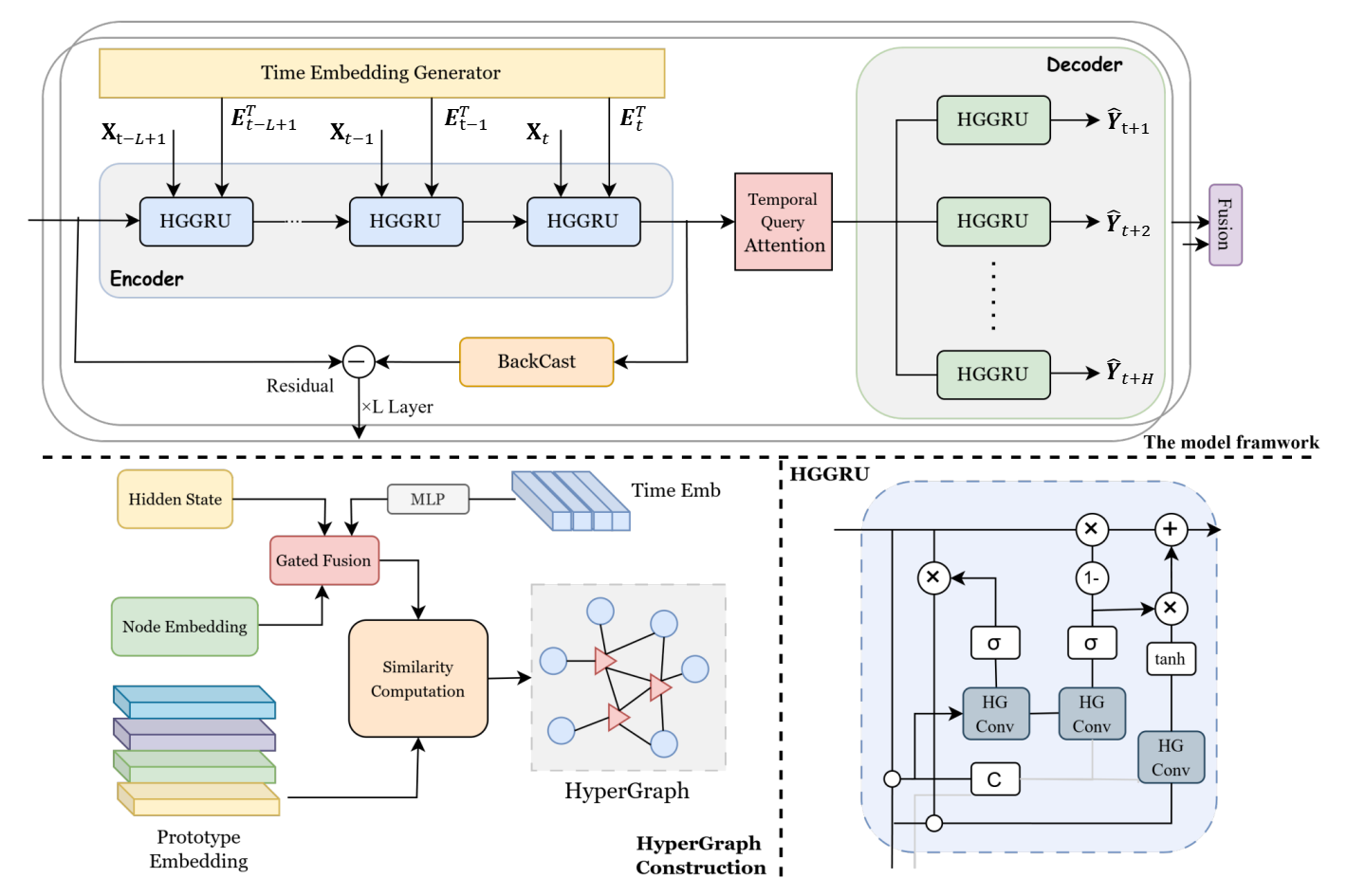}
    \caption{The framework of PHGNet and detailed components. PHGNet is primarily composed of HGGRU modules, the structure of which is illustrated in the bottom-right panel. The HyperGraph Convolution (HGconv) module, shown in the bottom-left panel, details the process of hypergraph construction.}
    \label{fig:HeteroHGNet}
\end{figure*}

Despite these advances, traffic forecasting remains challenging due to a fundamental issue: \textbf{spatial heterogeneity}. As shown in Fig.\ref{fig:spatial hetero}, although nodes A and B are geographically adjacent and exhibit similar historical observations, their future traffic states differ significantly. This is mainly because nodes play inherently different functional roles~(e.g., residential or commercial areas), leading to distinct underlying dynamics. Such functional roles can be reflected in representative traffic patterns, and nodes sharing similar patterns often evolve in group-wise manner. As a result, spatial correlations driven by pattern similarity go beyond isolated pairwise edges, exhibit high-order relational dependencies at the regional level. However, existing methods have limitations facing this challenge. Distance-based graphs rely on predefined priors and miss this nearby-but-dissimilar phenomenon. Other learnable graph methods are restricted to pairwise relationship modeling, limiting their ability to capture high-order interactions. Moreover, Fig.\ref{fig:spatial hetero} further reveals a remarkable difference in node correlations between various time steps~(Red and green box). This suggests that high-order spatial correlations also evolve over time, making it difficult to model such relations in a stable and dynamic manner.

To address the above challenges, we propose PHGNet, a unified spatiotemporal forecasting framework based on \textbf{P}rototype-Guided \textbf{H}yper\textbf{G}raph construction for modeling heterogeneous traffic dynamics. Overall, PHGNet adopts an Encoder–Decoder architecture, where Hypergraph Gated Recurrent Units (HGGRUs) serve as the core building block. Within each HGGRU step, nodes are dynamically assigned to learnable prototype embeddings to form hyperedges, enabling the modeling of adaptive high-order spatial interactions among nodes with similar traffic patterns. Furthermore, to improve the robustness of dynamic hypergraph construction against short-term noise and fluctuations, we introduce a global–local node representation that fuses static spatial priors with dynamic hidden states. This design enables dynamic hypergraph construction while preserving spatiotemporal consistency. For forecasting, we introduce an iterative residual learning strategy to hierarchically refine predictions. The Temporal Query Attention is further developed for parallel decoding, thereby improving both forecasting accuracy and inference efficiency.

Our main contributions are summarized as follows.

\textbullet\ We propose PHGNet, a prototype-guided hypergraph framework for traffic forecasting, with HGGRU serving as its core building block. The model can capture high-order and dynamic spatial dependencies to effectively alleviate the challenges of spatial heterogeneity.

\textbullet\ We design an Adaptive Hypergraph Convolution module where hypergraph construction is supported by global-local node representation and prototype-based assignment.  This module ensures both robustness and dynamics in hypergraph message propagation.

\textbullet \ We conduct extensive experiments on real-world traffic datasets to evaluate the performance of PHGNet. The results demonstrate that our model outperforms state-of-the-art ones, while the integration of iterative residual learning and parallel decoding further improves forecasting accuracy and efficiency.

\section{Methodology}
In this section, we introduce the proposed PHGNet framework, as illustrated in Fig. \ref{fig:HeteroHGNet}. The model comprises three main components and the details are presented in this section.

\subsection{Spatio-Temporal Embedding Module}
This module is designed to encode essential prior information, enabling the model to capture complex spatiotemporal dynamics in traffic flow.

Specifically, to model periodic temporal patterns of traffic states,, we learn two temporal embeddings: time-of-day (ToD) embedding and day-of-week (DoW) embedding. Given the past $L$ time step tod indices $T^d \in \{1, \dots, 288\}^L$ and dow indices $T^w \in \{1, \dots, 7\}^L$~(288 slots per day under 5-minute sampling and 7 days per week), we first encode them into one-hot vectors and then project them into embedding spaces to obtain $\mathbf{E}^d$ and $\mathbf{E}^w \in \mathbb{R}^{L \times D_T}$ through bias-free linear layers, where $D_T$ is the dimension of time embedding. Both of these are collectively denoted as $\mathbf{E}^T$ in the figures.

Moreover, we introduce a learnable spatial embedding $\mathbf{E}^s \in \mathbb{R}^{N \times D_N}$ to adaptively model the global characteristics of each node, where $N$ is the number of nodes, and $D_N$ is the spatial embedding dimension. The time and node embeddings will be utilized as key inputs for the subsequent stages of the model.

\subsection{Prototype-based HyperGraph Convolution}
After obtaining the spatio and temporal embeddings, we introduce hypergraph to address the limitation of spatial heterogeneity. Specifically, we propose a prototype-based hypergraph convolution module consisting of two components: (1) hypergraph construction and (2) hypergraph convolution.

\subsubsection{Global-Local Node Representation}
A critical step in hypergraph construction is to derive reliable node representations for structure inference. Such representations should be both time-adaptive and temporally consistent: relying solely on local features may introduce short-term noise, leading unstable hypergraph structure, while purely using static global priors cannot adapt to distribution shifts in traffic patterns.

Therefore, we propose a global–local node representation module as the basis for hypergraph construction. In detail, we fuse the local input hidden state $\textbf{H}_{t-1} \in \mathbb{R}^{N \times D_h}$ with the global static spatial embedding $\textbf{E}^s$ obtained before. The fusion is modulated by a time-conditioned gate, and the resulting node representation $\mathbf{E}_t^N$ at time step $t$ can be formulated as follows:
\begin{equation}
\mathbf{E}_t^N=\alpha_t\,\mathbf{H}_{t-1}\mathbf{W}_h+\big(1-\alpha_t\big)\mathbf{E}^s,
\label{eq:global_local}
\end{equation}
Here $\alpha_t = \sigma((\mathbf{E}^d_t \odot \mathbf{E}^w_t)\mathbf{W}_T) \in (0,1)$, where $\sigma$ denotes the sigmoid function. $\mathbf{W}_h \in \mathbb{R}^{D_h \times D_N}$ and $\mathbf{W}_T \in \mathbb{R}^{D_T \times D_N}$ are both learnable parameters to align the embedding dimension.

\subsubsection{HyperGraph Construction}
Based on the obtained node representations, We construct a prototype-driven hypergraph. Unlike traditional methods that rely on fixed graph topology, our approach utilizes a learnable prototype bank $\mathbf{P} = [\mathbf{P}_1, \ldots, \mathbf{P}_M] \in \mathbb{R}^{M \times D_N}$, where each prototype encodes a latent traffic pattern. Based on this, the assignment matrix $\mathbf{S}_t \in \mathbb{R}^{N \times M}$ at time step $t$ is computed as:
\begin{equation}
\mathbf{S}_t = \mathrm{softmax}\!\left(\mathbf{E}_t^N\mathbf{P}^\top\right),
\label{eq:n2e}
\end{equation}
Here, $\mathbf{S}_t$ provides a soft assignment of nodes to prototypes, allowing nodes to adaptively select the most relevant patterns under different traffic conditions.

\subsubsection{HyperGraph Convolution}
After obtaining the prototype-based assignment matrix $\mathbf{S}_t$, we perform hypergraph convolution to propagate high-order spatial information. The convolution consists of two stages: node-to-hyperedge aggregation and hyperedge-to-node aggregation.
In the process of node-to-hyperedge aggregation, node features are aggregated into hyperedge representations with degree normalization. The resulting hyperedge features, which capture representative traffic patterns, are computed as follows:
\begin{equation}
\mathbf{H}_t^e = \mathbf{S}^{\top}_t\left(\mathbf{D}_{v,t}^{-\frac{1}{2}} \mathbf{X}^{E}_t\right) \mathbf{W}_e .
\label{eq:n3e}
\end{equation}
where $\mathbf{X}_t^E \in \mathbb{R}^{ N \times D}$ is a $D$-dimensional embedding of input $\mathbf{X}$ at time step $t$. $\mathbf{D}_v \in \mathbb{R}^{ N \times N}$ is the diagonal node-degree matrix, and $\mathbf{W}_e \in \mathbb{R}^{D \times D}$ is a learnable matrix.

In the hyperedge-to-node aggregation stage, hyperedge features are propagated back to nodes with both hyperedge-degree and node-degree normalization. The updated node representations are obtained as:
\begin{equation}
\mathbf{X}^h_t=\mathrm{ReLU}\left(\mathbf{D}_{v,t}^{-\frac{1}{2}}\mathbf{S}_t\big(\mathbf{D}_{e,t}^{-1} \mathbf{H}_t^e\big)\right),
\label{eq:e4n}
\end{equation}
where $\mathbf{D}_e \in \mathbb{R}^{M \times M}$ denotes a diagonal hyperedge-degree matrix. 
This two-stage message passing process enables nodes to receive high-order information through adaptive hyperedges.
Finally, we concatenate the hypergraph convolution output $\mathbf{X}^h_t$ with the embedded inputs $\mathbf{X}^{E}_t$ to perform 2-order feature transformation:
\begin{equation}
\mathbf{H}^o_t = [\mathbf{X}^h_t||\mathbf{X}^E_t]\mathbf{W}_o=[\mathbf{X}^h_t||\mathbf{X}^E_t]\mathbf{E}^N_t
\mathbf{\Theta}
\label{eq:output}
\end{equation}
where a node-adaptive parameter learning mechanism is further used to enhance expressiveness. The parameter $\mathbf{W}_o$ is dynamically generated by the product of above global–local node representation $\mathbf{E}^N_t$ and a $2$-order weight pool $\
\mathbf{\Theta} \in \mathbb{R}^{D_N \times 2 \times D \times D_{h}}$, allowing the model to capture unique traffic patterns for each node.

\subsection{Encoder-Decoder Framework}
To jointly capture temporal dynamics and high-order spatial dependencies within a unified framework, our model adopts an Encoder–Decoder architecture built upon HGGRU with iterative residual learning. Specifically, HGGRU extends the standard GRU by replacing linear transformations with the proposed prototype-based hypergraph convolution, as follows:
\begin{equation}
\begin{aligned}
\mathbf{r}_t &= \sigma\!\left(\mathrm{HGConv}\left([\mathbf{X}_t \,\Vert\, \mathbf{H}_{t-1}]; \vartheta_r\right)\right), \\
\mathbf{u}_t &= \sigma\!\left(\mathrm{HGConv}\left([\mathbf{X}_t \,\Vert\, \mathbf{H}_{t-1}]; \vartheta_u\right)\right), \\
\tilde{\mathbf{H}}_t &= \tanh\!\left(\mathrm{HGConv}\left([\mathbf{X}_t \,\Vert\, (\mathbf{r}_t \odot \mathbf{H}_{t-1})]; \vartheta_h\right)\right), \\
\mathbf{H}_t &= \mathbf{u}_t \odot \mathbf{H}_{t-1} + (1-\mathbf{u}_t)\odot \tilde{\mathbf{H}}_t .
\end{aligned}
\end{equation}
where $\mathbf{X}_t$ and $\mathbf{H}_t$ are the input and hidden state at times step $t$, respectively. $\odot$ denotes the hardmard product. $\mathrm{HGConv}(\cdot)$ denotes the prototype-based HyperGraph Convolution proposed before. The terms $\mathbf{r}_t$ and $\mathbf{u}_t$ are reset and update gate respectively, $\vartheta_{r}$, $\vartheta_{u}$ and $\vartheta_{h}$ are parameters generated by the above module.

To mitigate the distribution shift between historical inputs and future outputs, we introduce a Temporal Query Attention to bridge the Encoder and Decoder. Specifically, it employs time-conditioned cross-attention to retrieve relevant historical patterns for each forecasting horizon, where the future $H$-step time embeddings $\mathbf{E}^T_f = \mathbf{E}^d_f \odot \mathbf{E}^w_f$ are used as queries, the historical time-aware  representations formed by the concatenation of past time embeddings $\mathbf{E}^T_p = \mathbf{E}^d_p \odot \mathbf{E}^w_p$ and hidden output $\mathbf{H}_p$ are used to form key and value. The computation process of the above module is formulated as follows:
\begin{equation}
\begin{aligned}
\mathbf{Q} &= \mathbf{E}^T_f \mathbf{W}_{Q},\\
\mathbf{K} &= (\mathbf{E}^T_p || \mathbf{H}_p)  \mathbf{W}_{K},\\
\mathbf{V} &= (\mathbf{E}^T_p || \mathbf{H}_p) \mathbf{W}_{V}.
\end{aligned}
\end{equation}
where $\mathbf{W}_Q, \mathbf{W}_K, \mathbf{W}_V$ are learnable parameters. Furthermore, inspired by PM-DMNet\cite{PM-DMNet}, we adopt a parallel decoder for efficient multi-step forecasting. Specifically, $H$ parallel decoder states are initialized, along with the outputs from the Temporal Query Attention, are fed into a HGGRU block to directly generate the future output $\hat{\mathbf{Y}}$ in a single forward pass.

To further refine the predictions, we employ a residual learning mechanism consisting of $L$ stacked blocks. In the $i$-th block, the encoder hidden states are used to generate a forecast component $\hat{\mathbf{Y}}_i$ and a backcast component $\hat{\mathbf{X}}_i$. Here, $\hat{\mathbf{X}}_i$ can be explained as the learned part of the original input, while the residual $\mathbf{X} - \hat{\mathbf{X}}_i$ is passed to the next block for further modeling. The final prediction is obtained by summation of all forecasting blocks and can be formulated as $\hat{\mathbf{Y}} = \sum_{i=1}^L\hat{\mathbf{Y}} _i$.

\subsection{Training Strategy}
To optimize the model, the Mean Absolute Error (MAE) is employed as the loss function, which is formulated as follows:
\begin{equation}\label{eq:mae_loss}
\mathcal{L}=\frac{1}{H \cdot N}\sum_{i=1}^{H}\sum_{j=1}^{N}\left|\mathbf{Y}^{gt}_{i,j}-\hat{\mathbf{Y}}_{i,j}\right|.
\end{equation}
where $\mathbf{Y}^{gt}_{i,j}$ denotes the element in the i-th row and j-th column of the ground truth.

\section{Experiments}
\subsection{Experiment Settings}\label{AA}
\textbullet \ \noindent\textbf{\textit{Datasets:}} To assess the effectiveness of PHGNet, we conducted experiments on four benchmark traffic datasets: PeMS03, PeMS04, PeMS07, and PeMS08, which are sourced from California Performance Measurement System (PeMS). The details of them are summarized in Table \ref{tab:datasets}. Each of them maintains a 5-minute sampling frequency, corresponding to 12 observations per hour.

\begin{table}[!b]
\caption{Datasets Description.}
\begin{center}
\begin{tabular}{|c|c|c|c|}
\hline
\textbf{Datasets} & \textbf{Node} & \textbf{Time step} & \textbf{Time Range} \\
\hline
PeMS03 & 358 & 26202 & 09/01/2018--11/30/2018 \\
\hline
PeMS04 & 307 & 16992 & 01/01/2018--02/28/2018 \\
\hline
PeMS07 & 883 & 28224 & 05/01/2017--08/31/2017 \\
\hline
PeMS08 & 170 & 17856 & 07/01/2016--08/31/2016 \\
\hline
\end{tabular}
\label{tab:datasets}
\end{center}
\end{table}

\begin{table*}[t]
\caption{Performance Comparison on PEMS datasets.}
\begin{center}
\scriptsize
\renewcommand{\arraystretch}{1.15}
\setlength{\tabcolsep}{3.5pt}

\resizebox{\textwidth}{!}{%
\begin{tabular}{|c|ccc|ccc|ccc|ccc|}
\hline
\textbf{Method} &
\multicolumn{3}{c|}{\textbf{PEMS03}} &
\multicolumn{3}{c|}{\textbf{PEMS04}} &
\multicolumn{3}{c|}{\textbf{PEMS07}} &
\multicolumn{3}{c|}{\textbf{PEMS08}} \\
\cline{2-13}
 & \textbf{MAE} & \textbf{RMSE} & \textbf{MAPE} &
   \textbf{MAE} & \textbf{RMSE} & \textbf{MAPE} &
   \textbf{MAE} & \textbf{RMSE} & \textbf{MAPE} &
   \textbf{MAE} & \textbf{RMSE} & \textbf{MAPE} \\
\hline

\hline
ARIMA\cite{ARIMA} & 35.41 & 47.59 & 33.78\% & 33.73  & 48.80 & 24.18\% & 38.17 & 59.27 & 19.46\% & 31.09 & 44.32 & 22.73\% \\
STGCN\cite{STGCN}  & 17.49 & 30.12 & 17.15\% & 22.70 & 35.55 & 14.59\% & 25.38 & 38.78 & 11.08\% & 18.02 & 27.83 & 11.40\% \\
DCRNN\cite{DCRNN}  & 18.18 & 30.31 & 18.91\% & 24.70 & 38.12 & 17.12\% & 25.30 & 38.58 & 11.66\% & 16.82 & 26.36 & 10.92\% \\
GWNet\cite{GWNet}  & 19.85 & 32.94 & 19.31\% & 25.45 & 39.70 & 17.29\% & 26.85 & 42.78 & 12.12\% & 19.13 & 31.05 & 12.68\% \\
GMAN\cite{GMAN} & 16.87 & 27.92 & 18.23\% & 19.14 & 31.60 & 13.19\% & 20.97 & 34.10 & 9.05\% & 15.31 & 24.92 & 10.13\% \\
DGCRN\cite{DGCRN}  & \underline{14.74} & \underline{25.97} & 15.42\% & 18.73 & 30.65 & 12.82\% & 20.34 & \underline{33.22} & 8.84\% & 14.30 & \underline{23.50} &  \underline{9.33}\% \\
D$^2$STGNN\cite{D2STGNN} & 15.10 & 26.57 & 15.23\% & \underline{18.42} & \textbf{29.97} & 12.81\% & \underline{19.68} & 33.24 & 8.43\% & 14.35 & 24.18 & \underline{9.33}\% \\
LASTGNN\cite{LASTGNN} & 15.41 & 27.21 & 14.93\% & 19.55 & 31.88 & 12.70\% & 21.03 & 34.56 &  8.53\% & 15.63 & 24.78 &  9.78\% \\
DMGSTCN\cite{DMGSTCN} & 15.40 & 27.02 & \textbf{14.25}\% & 18.45 &  \underline{30.26} & \underline{12.38}\% & 20.22 & 33.49 & \underline{8.19}\% & \underline{14.22} & 23.52 & 9.76\% \\
STDN\cite{STDN}  & 16.05 & 27.51 & 17.71\% & 18.67 & 30.92 & 13.16\% & 22.94 & 36.06 & 10.32\% & 14.79 & 24.60 & 10.26\% \\
\hline

\textbf{PHGNet} &
\textbf{14.49} & \textbf{25.58} & \underline{14.59}\% &
\textbf{18.17} & 30.58 & \textbf{12.01}\% &
\textbf{19.16} & \textbf{32.96} & \textbf{8.10}\% &
\textbf{13.40} & \textbf{23.18} & \textbf{8.88}\% \\
\hline
\end{tabular}%
}
\label{tab:pems_main_noLeftCol}
\end{center}
\end{table*}

\textbullet \ \noindent\textbf{\textit{Baselines:}} To assess the performance of PHGNet, We selected 10 baselines for traffic prediction, including traditional method: ARIMA \cite{ARIMA}, traditional deep learning methods: STGCN\cite{STGCN}, DCRNN\cite{DCRNN}, GWNet\cite{GWNet}, GMAN\cite{GMAN}, DGCRN\cite{DGCRN} and recent SOTA methods: D$^2$STGNN\cite{D2STGNN}, LASTGNN\cite{LASTGNN}, DMGSTCN\cite{DMGSTCN} and STDN\cite{STDN}.

\textbullet \ \noindent\textbf{\textit{Evaluation Metrics:}} We quantify the performance of methods on three metrics, including Mean Absolute Error~(MAE), Root Mean Squared Error~(RMSE), and Mean Absolute Percentage Error~(MAPE).

\textbullet \ \noindent\textbf{\textit{Parameter Settings:}} Our experiments are conducted on the NVIDIA RTX 4090 GPUs running CUDA 12.1. Each dataset is divided into training, validation, and test sets with a 6:2:2 ratio. We use historical data from the past 12 time steps to predict the next 12 time steps, and compare the average metric. The models are optimized using Adam with an initial learning rate of 0.001. To prevent overfitting, an early-stopping strategy is employed. The default batch size is set to 64, and is adaptively scaled to 32 or 16 in case of GPU memory constraints. For all datasets, the input dimension $C = 3$.

Totally, there are 3 key hyperparameters in our model: the number of prototype embeddings $M$, the hidden dimension of temporal embeddings $D_T$, and the hidden dimension of node embeddings $D_N$. During our experiments, $M$ is selected from $\{4, 6, 8, 10\}$, while both $D_T$ and $D_N$ are tuned within the range of $\{10, 12, 15, 20, 24, 36\}$.

\subsection{Performance Comparison}
The overall performance comparison is summarized in Table \ref{tab:pems_main_noLeftCol}. For all comparisons, the best results are highlighted in bold, while the second-best are underlined. From these results, we can make the following observations:

1) Overall, our proposed PHGNet achieves the best performance on most datasets and metrics, demonstrating consistent improvements over all compared methods. Notably, it achieves 5.76\%, 1.50\%, and 4.82\% relative improvements in MAE, RMSE, and MAPE respectively.
2) Traditional statistical methods generally perform poorly compared to other methods. This performance gap mainly because they primarily consider temporal correlations while neglecting spatial dependencies.
3) Graph-based models, such as GWNet and DCRNN, have achieved significant performance gain by jointly modeling spatial and temporal dependencies. Moreover, methods using learnable graph often surpass those using predefined graph. However, most of these models still rely on static graph construction, making it difficult to capture the dynamic nature of spatial relationships. 
4) Methods like DGCRN and DMGSTCN further enhance performance by modeling dynamic spatial dependencies, thereby outperforming most baselines. Nevertheless, they still lack sufficient consideration of high-order spatial correlations. Compared with these baselines, our proposed PHGNet models high-order spatial dependencies via prototype-guided hypergraphs and stabilizes dynamic structure learning with global–local node representations, consistently achieving superior performance.

\begin{table}
\caption{Ablation Study on PeMS07 and PeMS08}
\begin{center}
\scriptsize
\renewcommand{\arraystretch}{1.15}
\setlength{\tabcolsep}{3.5pt}

\begin{tabular}{|c|ccc|ccc|}
\hline
\textbf{Method} &
\multicolumn{3}{c|}{\textbf{PEMS07}} &
\multicolumn{3}{c|}{\textbf{PEMS08}} \\
\cline{2-7}
 & \textbf{MAE} & \textbf{RMSE} & \textbf{MAPE} &
   \textbf{MAE} & \textbf{RMSE} & \textbf{MAPE} \\
\hline
w/o Res & \underline{19.29} & 33.19 & \underline{8.16}\% & \underline{13.43} & \underline{23.44} & \underline{8.92}\% \\
w/o Res+Local & 19.46 & 33.40 & 8.18\% & 13.68 & 23.74 & 9.07\% \\
w/o Res+Global & 19.68 & 33.10 & 8.61\% & 13.61 & 23.51 & 8.98\% \\
w/o Res+DGC & 19.65 & 33.71 & 8.22\% & 13.91 & 23.97 & 9.17\% \\
w/o TE & 19.78 & 33.62 & 8.35\% & 15.06 & 24.51 & 9.68\% \\
w/o NAPL & 19.55 & \textbf{32.90} & 8.23\% & 13.75 & 23.45 & 9.06\% \\
\hline
\textbf{PHGNet} &
\textbf{19.16} & \underline{32.96} & \textbf{8.10}\% &
\textbf{13.40} & \textbf{23.18} & \textbf{8.88}\% \\
\hline
\end{tabular}

\label{tab:ablation}
\end{center}
\end{table}

\subsection{Ablation Study}
To analyze the contribution of each component in PHGNet, we conduct extensive ablation studies by comparing the full model with six variants.
1) w/o Res: Remove the residual learning structure. 
2) w/o Res+Local: Based on w/o Res, replace the global-local node representation with only local input embedding. 
3) w/o Res+Global: Based on w/o Res, replace the global-local node representation with only global static embedding.
4) w/o Res+DGC: Based on w/o Res, replace the hypergraph structure with adaptive graph convolution. 
5) w/o TE: remove the time embedding. 
6) w/o NAPL: Remove Node-adaptive parameter learning mechanism.

The results of ablation study on PeMS07 and PeMS08 are reported in Table \ref{tab:ablation}. Several key observations can be derived.
1) w/o Res underperforms the full model, indicating that residual framework can effectively compensate for information loss. However, this variant still outperforms most baselines, further validating the robustness of our HGGRU module. 
2) Both w/o Res+Local and w/o Res+Global exhibit a performance drop. suggesting that the integration of local and global embeddings not only stabilizes the training but also enhances the model's capacity to represent consistent spatiotemporal dependencies. 
3) w/o Res+DGC shows inferior results compared to hypergraph-based variants, demonstrating that our hypergraph architecture is more effective to capture high-order spatial relationships and mitigate spatial heterogeneity.
4) w/o TE exhibits the largest performance drop, confirming the importance of time embeddings in modeling time-varying dynamics in traffic data. 
5) w/o NAPL shows that adaptive node parameters effectively increase the model expressiveness.

\subsection{Parameter Sensitivity Analysis}
\begin{figure}[htbp]
    \centering
    \includegraphics[width=\columnwidth]{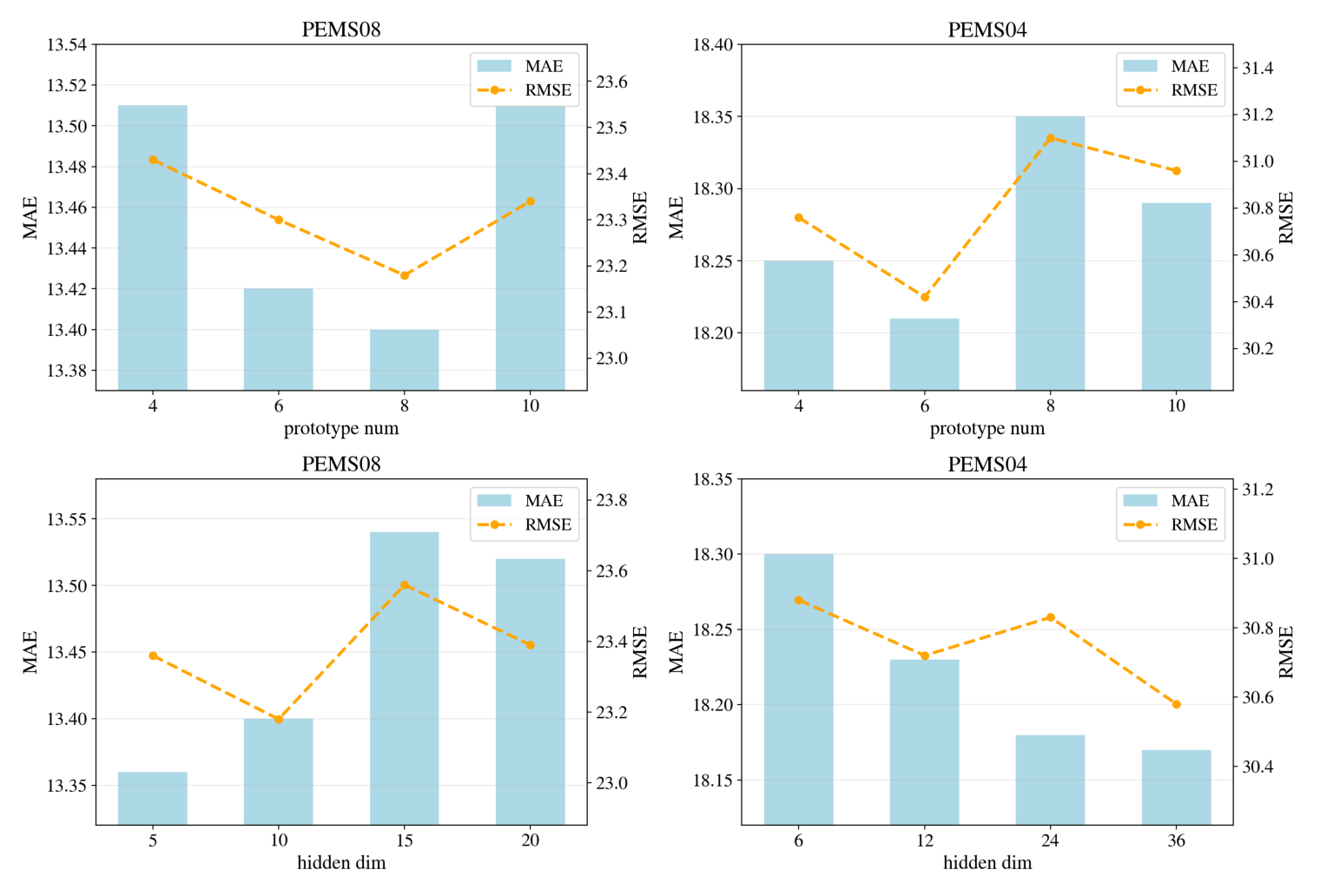} 
    \caption{Parameter senesitivity analysis on PeMS04 and PeMS08.}
    \label{fig:sensitivity}
\end{figure}
Fig. \ref{fig:sensitivity} reports the parameter sensitivity results on PeMS04 and PeMS08. In experiment, we investigate two key parameters: the number of prototype embeddings $M$ and the hidden dimension of node embeddings $D_N$. Specifically, $M$ is evaluated over $\{4,6,8,10\}$. For $D_N$, we test $\{5,10,15,20\}$ on PeMS08 and $\{6,12,24,36\}$ on PeMS04.

According to the results, we draw the following two observations: 
1) The model achieves the best performance with $M=8$ on PeMS08 and $M=6$ on PeMS04, indicating that a moderate prototype size can make more accurate prediction, while further increasing $M$ may introduce overfitting. 
2) On PeMS08, a medium embedding dimension $D_N=10$ performs best. Since PeMS08 is relative small, larger dimension tends to hurt generalization and reduce performance. In contrast, on the more complex PeMS04, increasing $D_N$ consistently enhances the model's performance, with the optimal setting at $D_N=36$.

\begin{figure*}[t]
    \centering
    \includegraphics[width=1\textwidth]
    {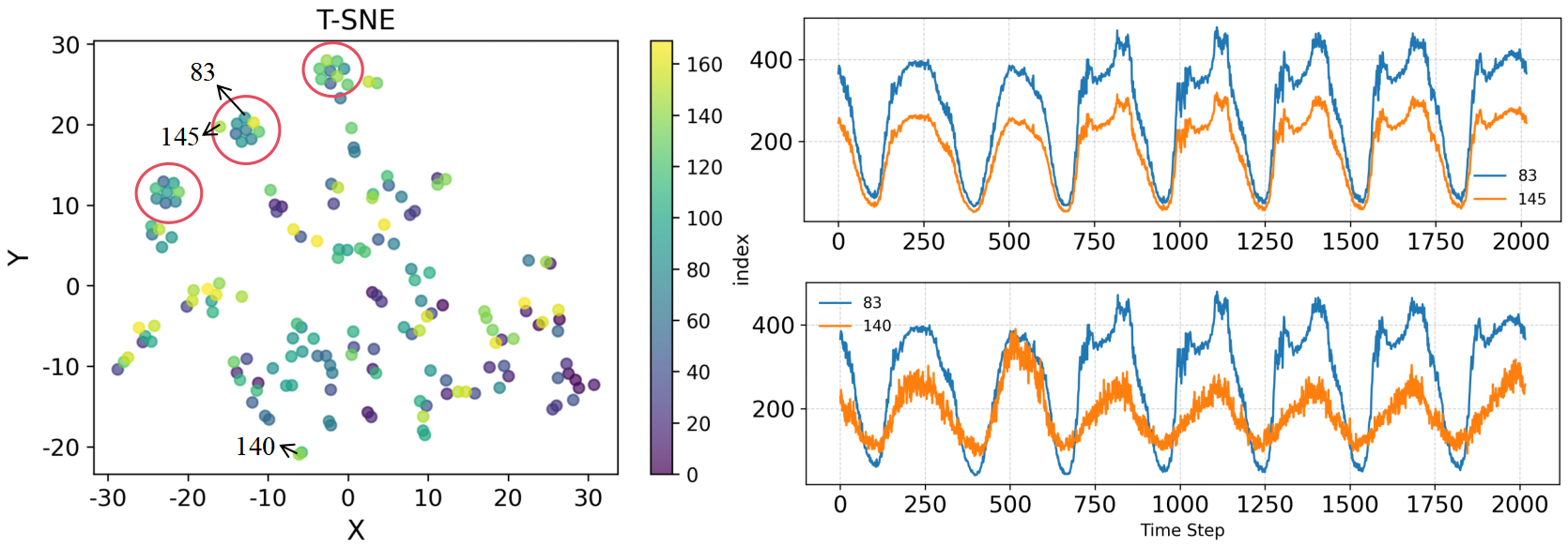} 
    \caption{T-SNE visualization of spatial embeddings on PeMS08 and corresponding traffic pattern similarity.}
    \label{fig:tsne}
\end{figure*}
\subsection{Visualization}
In this section, we present visualizations of the learned spatial embeddings and the corresponding traffic patterns. Specifically, we utilize T-SNE to project the learned spatial embeddings into a 2D space with a perplexity of 24. As depicted in Fig. \ref{fig:tsne}, the learned spatial embeddings are not randomly distributed; instead, they form several clear clusters, as highlighted by the red circles, indicating that nodes with similar traffic patterns are mapped to nearby regions. Meanwhile, some nodes remain more dispersed, reflecting diverse and heterogeneous traffic behaviors across the road network.

Furthermore, we compare the traffic flows of representative nodes in embedding space. For example, Nodes 83 and 145, which are close in the embedding space, exhibit similar temporal patterns with consistent peaks and trends. In contrast, node 140 is far from node 83 in the embedding space and shows significantly different temporal dynamics. These observations further validate the effectiveness of the proposed method. 

\section{Conclusion}
In this paper, we propose PHGNet, a novel Prototype-Guided Hypergraph Network designed to address the fundamental challenge of spatial heterogeneity in traffic forecasting. Compared with previous studies that model spatial structures through pairwise relationships, PHGNet leverages the prototype learning mechanism to adaptively construct hypergraph structure, effectively capturing dynamic high-order pattern similarities. A global–local node representation is also introduced to ensure spatiotemporal consistency. In addition, an iterative residual learning strategy and a parallel decoding are adopted to improve the forecasting accuracy and inference efficiency. Extensive experiments on four real-world traffic flow datasets demonstrate that our proposed PHGNet consistently reduces the MAE, RMSE and MAPE by 5.76\%, 1.50\% and 4.82\%, respectively, compared to state-of-the-art baselines.


\section*{Acknowledgment}
This work is supported by Shenzhen Ubiquitous Data Enabling Key Lab under grant ZDSYS20220527171406015, and by Tsinghua Shenzhen International Graduate School-Shenzhen Pengrui Endowed Professorship Scheme of Shenzhen Pengrui Foundation.

\bibliographystyle{IEEEtran} 
\bibliography{references}    

\end{document}